\documentclass[letterpaper]{article} 
\usepackage{aaai2026}  
\usepackage{times}  
\usepackage{helvet}  
\usepackage{courier}  
\usepackage[hyphens]{url}  
\usepackage{graphicx} 
\usepackage{natbib}  
\usepackage{caption} 
\usepackage[ruled,vlined,linesnumbered]{algorithm2e}    

\usepackage{amsmath,amssymb}
\usepackage{mathrsfs}

\usepackage{pgfplots}
\usepgfplotslibrary{fillbetween}
\pgfplotsset{compat=1.18}
\usetikzlibrary{automata, positioning, arrows.meta, bending}
\usepackage{subcaption}

\usepackage{stix}
\DeclareMathAlphabet\mathbfcal{LS2}{stixcal}{b}{n}
\usepackage{hyperref}

\definecolor{expcolor0}{RGB}{76,114,176}
\definecolor{expcolor1}{RGB}{221,132,82}
\definecolor{expcolor2}{RGB}{85,168,104}
\definecolor{expcolor3}{RGB}{196,78,82}

\nocopyright{}

\title{Pareto Q-Learning with Reward Machines}
\author{
    Arnaud Lequen\textsuperscript{1}\equalcontrib,
    Clément Legrand-Lixon\textsuperscript{2}\equalcontrib,
    Léo Saulières\textsuperscript{3}\equalcontrib
}
\affiliations{
    \textsuperscript{1}Linköping University, Sweden\\
    \textsuperscript{2}Univ. Lille, CNRS, Centrale Lille, UMR 9189 CRIStAL, F-59000 Lille, France\\
    \textsuperscript{3}Univ. Toulouse, INRAE-MIAT, Toulouse, France\\
    arnaud.lequen@liu.se, clement.legrand-lixon@univ-lille.fr, leo.saulieres@inrae.fr
}

\newcommand{\stateset}{\ensuremath{\mathcal{S}}}
\newcommand{\actionset}{\ensuremath{\mathcal{A}}}

\newcommand{\probfunc}{\ensuremath{p}}
\newcommand{\ssucc}{\ensuremath{\mathsf{succ}}}

\newcommand{\policy}{\ensuremath{\pi}}

\renewcommand{\rm}{\ensuremath{\mathcal{R}}}
\newcommand{\rmvec}{\ensuremath{\mathbfcal{R}}}
\newcommand{\statetf}{\ensuremath{\delta_{\textit{st}}}}
\newcommand{\rewardtf}{\ensuremath{\delta_{\textit{re}}}}

\definecolor{nicegreen}{HTML}{72BE79}
\definecolor{nicered}{HTML}{DE6E59}

\begin{document}

\maketitle

\begin{abstract}
We present Pareto Q-Learning with Reward Machines (PQLRM), a multi-objective reinforcement learning algorithm for tasks whose reward structure is specified by a set of reward machines (RMs). PQLRM combines Pareto Q-Learning (PQL), which maintains sets of vector-valued Q-estimates to approximate the Pareto front, with enhancements from Q-Learning with Reward Machines (QRM), which exploits the factored automaton structure of the reward signal. This yields a multi-policy algorithm that remains sample-efficient under non-Markovian, RM-encoded rewards. Experimental trials show that PQLRM converges faster than a naive PQL baseline applied to the cross-product MDP and can synthesize Pareto-optimal policies that QRM cannot.
\end{abstract}

\section{Introduction}

Reinforcement Learning (RL) is a paradigm for sequential decision-making in which an agent learns to act through interaction with an environment, with the aim of maximizing a cumulative reward signal. It has driven significant breakthroughs across a broad range of domains, including video games~\cite{badia2020agent57}, traffic control~\cite{kusic2021spatial}, computer networks~\cite{omoniwa2022energy}, and robotics~\cite{gurtler2023real}. Central to RL is the design of a reward function that encodes the desired behaviour of the agent. Yet, real-world tasks rarely reduce to a single scalar objective: they typically demand balancing several, often conflicting criteria such as performance, safety, or energy consumption.

This observation has motivated the development of Multi-Objective Reinforcement Learning (MORL), which extends the classical RL framework to vector-valued rewards, each component corresponding to a distinct objective. In such cases, no single policy is universally optimal. Instead, MORL seeks to approximate the \emph{Pareto front}, i.e., the set of policies that are non-dominated across objectives, so that a user can later select an appropriate trade-off. Efficiently learning such a set, however, remains challenging, especially when the reward structure of the task is itself complex.

Reward Machines (RMs)~\cite{toroicarte2018using} have recently emerged as a powerful abstraction for representing non-Markovian reward functions. A reward machine is a finite-state automaton that encodes the reward structure of a task in a modular and structured way. This representation offers several advantages. First, it provides an interpretable description of the reward function, making task specifications easier to design, understand and verify. Second, the explicit state structure of reward machines can be exploited to accelerate learning, for example by decomposing value functions across the states of the automaton. 

In this work, we bring together the strengths of reward machines and multi-objective reinforcement learning. We propose a novel multi-objective RL algorithm that integrates reward machines with Pareto Q-Learning~\cite{moffaert2014multi} to learn a set of non-dominated policies. Our approach leverages the structural decomposition induced by reward machines to guide exploration and value propagation across objectives, while maintaining a Pareto-based representation of value functions. As a result, the algorithm is inherently multi-policy, as it synthesizes a set of trade-off solutions after a single run.


\section{Background}



\subsection{Multi-Objective Reinforcement Learning}

\paragraph{Multi-Objective Markov Decision Problem}
A Multi-Objective Markov Decision Problem (MOMDP)~\cite{4220828} is a tuple $M = \langle \stateset{}, \actionset, \mathbf{R}, \probfunc{}, \gamma \rangle$, where $\stateset{}$ and $\actionset$ denote, respectively, the state space and action space, and $\gamma \in (0, 1]$ is the discount factor.
$\mathbf{R} : \stateset{} \times \actionset \times \stateset{} \rightarrow \mathbb{R}^d$ is the  \emph{reward function}, which is a function that returns a vector composed of $d$ scalars, each representing the reward associated with an objective. 
$p : \mathcal{S} \times \mathcal{A} \times \mathcal{S} \rightarrow [0, 1]$ denotes the transition function, where, for all $s \in \stateset{}, a \in \actionset{}$, $p(s,a, \cdot)$ is a probability distribution over $s' \in \stateset{}$. $p(s,a,s')$ is the probability of reaching the state $s'$ by performing action $a$ from state $s$. In this paper, however, we focus on deterministic MDPs, where $p(s, a, s') \in \{0, 1\}$. We will write $\ssucc{}: \stateset{} \times \actionset{} \rightarrow \stateset{}$ the function such that $\ssucc{}(s, a) = s'$, where $s'$ is the only state such that $p(s, a, s') = 1$.


A solution to an MOMDP is a \emph{policy}, which is a mapping $\policy{}: \stateset^+ \rightarrow \actionset$ (where $\stateset^+$ are tuples of states of size at least 1) that associates the history of visited states with the action that the agent chooses. For a given state sequence $s_0, \ldots, s_t$, $\policy(s_0, \ldots, s_t) = a_t$ is the action chosen by the agent at step $t$, leading to state $\ssucc(s_t, a_t) = s_{t+1}$.

\paragraph{Pareto-Dominance}
Let $\Pi$ be the set of all policies for an MOMDP $M$, and let $\policy, \policy' \in \Pi$. Policies are compared using their \emph{value functions} $\mathbf{V}^{\pi}$, defined as:
%
\[
    \mathbf{V}^{\pi} = \sum_{t=0}^{\infty} \gamma^{t}\mathbf{r}^{\policy}_{t+1}
\]
where $\mathbf{r}^{\policy}_{t}$ is the reward obtained at time step $t$ after unfolding the policy $\policy{}$.

Let $\mathbf{V}^{\pi}_i$ be the $i$-th component of the value function of $\pi$.
Policy $\policy{}$ then dominates $\policy{}'$, noted $\mathbf{V}^{\pi} \succ_P \mathbf{V}^{\pi'}$, if we have that, for all $i \leq d$, $\mathbf{V}^{\pi}_i \ge \mathbf{V}^{\pi'}_i$, and for at least one $i \leq d$, $\mathbf{V}^{\pi}_i > \mathbf{V}^{\pi'}_i$.

The Pareto front of $M$ is then the set of non-dominated policies $PF(\Pi)$:
%
\[
PF(\Pi) = \{\pi \in \Pi \mid \nexists \pi' \in \Pi : \mathbf{V}^{\pi'} \succ_P \mathbf{V}^{\pi}\}
\]
%

The Pareto front contains the set of \emph{Pareto-optimal} policies that are not Pareto dominated in the sense of $\succ_P$, i.e., policies for which there is no other policy with equal or greater value for all objectives. 



Since policy quality in the multi-objective setting cannot be scalarized, the convergence of the Pareto front is often evaluated through alternative metrics. A widely used one is the hypervolume~\cite{bader2011hype}, which measures the volume of the region dominated by the front relative to a fixed reference point.

\subsection{MORL with Reward Machines}

\paragraph{Reward Machines} A Reward Machine (RM) is a finite state machine that represents a global structured reward signal, where each transition is associated with a reward function \cite{toroicarte2018using}. 
Given a set of propositional symbols $\mathcal{P}$, a state space $\stateset{}$ and action space $\actionset$, an RM is defined as a tuple $\rm{} = \langle U, u_o, \statetf, \rewardtf \rangle$.
$U$ and $u_o \in U$ denote, respectively, a finite set of states and an initial state.
$\statetf~:~U \times 2^{\mathcal{P}} \rightarrow U$ denotes the state-transition function.
$\statetf(u,\sigma)$ is the state reached after receiving truth assignment $\sigma \in 2^\mathcal{P}$ while being in state $u$. $\rewardtf : U \times U \rightarrow [\stateset{} \times \actionset \times \stateset{} \rightarrow \mathbb{R}]$ denotes the reward-transition function; $\rewardtf(u,u')$ is the reward function to use when transitioning from state $u$ to $u'$ of the RM.

\paragraph{MOMDP with RMs}
An MOMDP with RMs (MOMDPRM) over the set of propositions $\mathcal{P}$ is a tuple $M = \langle \stateset{}, \actionset, \rmvec{}, \probfunc{}, \gamma, L \rangle$
where $\stateset{}$, $\actionset$, $\probfunc{}$ and $\gamma$ are defined as in an MOMDP. $\rmvec{} = \langle \rm{}^1, \ldots, \rm{}^d \rangle$ is a vector of $d$ reward machines $\rm{}^i = \langle U^i, u^i_o, \statetf^i, \rewardtf^i \rangle$, for all $i \leq d$. $L: \stateset{} \rightarrow 2^{\mathcal{P}}$ is a \emph{labelling function} that maps every state of the MOMDPRM with a truth assignment of the set of propositions it is built upon.

\subsection{Algorithms}


Multi-Objective RL algorithms can be split into two families~\cite{hayes2022practical}, depending on whether the user's preferences over objectives are known a priori. When such preferences are available, \emph{single-policy} algorithms can scalarize the vector reward into a single signal and learn one policy aligned with these preferences, reducing the problem to a standard single-objective RL problem. When preferences are unknown, \emph{multi-policy} algorithms instead synthesize a set of policies realising different trade-offs between the objectives, with the aim of finding policies on the Pareto front.

\paragraph{Pareto Q-Learning}

Pareto Q-learning (PQL)~\cite{moffaert2014multi} extends Q-learning~\cite{watkins1992q} to the multi-objective setting by maintaining, for each state-action pair $(s,a)$, a set $\hat{Q}_{set}(s,a)$ of vector-valued Q-values corresponding to Pareto-dominating policies.


One of the main mechanisms behind PQL is that the algorithm learns the immediate and future components separately: $\overline{\mathscr{R}}(s,a)$ stores the running average of the observed immediate reward vector after taking action $a$ in state $s$. $\textit{ND}_t(s,a)$ stores the set of non-dominated vectors reachable from $s$ with action $a$ at time $t$:
\begin{equation*}
    \textit{ND}_t(s,a) = \textit{ND}\!\left(\bigcup_{a'} \hat{Q}_{set}(s',a')\right)
\end{equation*}
where the $\textit{ND}$ operator only preserves the non-dominated vectors of the set it is provided.
The $\hat{Q}_{set}$ is then reconstructed at run time via:
\begin{equation}
    \hat{Q}_{set}(s,a) \leftarrow \overline{\mathscr{R}}(s,a) \oplus \gamma\, \textit{ND}_t(s,a) \label{eq:pql_qsets}
\end{equation}
where $\oplus$ denotes the vector-set sum, defined for a vector $\mathbf{v}$ and a set of vectors $V$ as:
\[
    \mathbf{v} \oplus V = \left\{ \mathbf{v} + \mathbf{v}' \mid \mathbf{v}' \in V \right\}
\]

This decoupling removes the need for explicit vector correspondence, but also lets each component converge independently.


A specific Pareto-optimal policy is reconstructed at execution time by committing to a target vector in $\bigcup_{a_o \in \actionset{}} \hat{Q}_{set}(s_0, a_0)$ and, at each subsequent step, selecting the action whose set contains the vector consistent with the chosen one, using the immediate reward to take into account the difference induced by the action applied.

\paragraph{Q-Learning with Reward Machines}

Q-learning with Reward Machines (QRM)~\cite{toroicarte2018using} exploits the structure of an RM to decompose a (possibly non-Markovian) task into a set of Markovian sub-tasks, one for each RM state.
For each state $u \in U$ of the RM, QRM maintains a separate Q-function $q^u(s,a)$ that predicts the expected return obtained from the environment state $s$ assuming the RM is currently in state $u$.

During an episode, the current agent's RM state $u_p$ is tracked.
The decision rule selects actions using the Q-function associated with $u_p$,
i.e. $q^{u_p}$.

Having multiple Q-functions may make the learning process significantly slower.
The key idea is to update \emph{all} Q-functions of an RM in parallel from a single transition $(s,a,s')$.
For every RM state $u_j$, QRM uses the RM to synthesise the reward $r_j = \rewardtf(u_j, u_k)(s,a,s')$ that would have been received from $u_j$ (where $u_k = \statetf(u_j, L(s'))$ is the corresponding next RM state) and applies the standard update:
\begin{equation*}
    q^{u_j}(s,a) \xleftarrow{\alpha} r_j + \gamma \max_{a'} q^{u_k}(s',a')
\end{equation*}

As the RM acts as a deterministic model of the reward structure, every transition observed in the environment yields a valid training sample for every sub-task simultaneously, which considerably improves sample efficiency.
QRM is shown to converge to an optimal policy of the underlying MDPRM in the tabular setting.

\section{Pareto Q-Learning with Reward Machines}

Our approach to solving MOMDPRMs essentially consists in enhancing PQL with the core ideas of QRM, allowing the algorithm to leverage the factored decomposition of the state space induced by the reward machines.






\begin{algorithm}[ht]
\SetKwInOut{Input}{Input}
\SetKwInOut{Output}{Output}
\Input{MOMDPRM $M$, number of episodes $T$}
\Output{A set of Q-sets $Q_{set}^{\vec{\textsf{v}}}(s,a)$}
Initialize $Q_{set}^{\vec{\textsf{v}}}(s,a)$ as empty sets \nllabel{ln:init-qset} \\
\For{$T$ episodes\nllabel{ln:episodes}}{
    Initialize state $s$ of the environment \nllabel{ln:init-state}\\
    $\vec{\textsf{u}} := \langle \textsf{u}_1, \ldots, \textsf{u}_d \rangle \leftarrow \langle u_0^1, \ldots, u_0^d \rangle$ \nllabel{ln:init-rm}
    \\
    \While{$s$ is not terminal\nllabel{ln:while}}{
        Choose action $a$ using $s$ and $\vec{\textsf{u}}$ \nllabel{ln:select-action}\\
        $s' \leftarrow \ssucc{}(s, a)$ \nllabel{ln:step}\\
        \For{each $\vec{\textsf{v}} := \langle \textsf{v}_1, \ldots, \textsf{v}_d \rangle$ of RM states\nllabel{ln:loop-v}}{
        $\sigma \leftarrow L(s')$ \nllabel{ln:label}\\
        $\vec{\textsf{v}'} \leftarrow \langle \statetf^1(\textsf{v}_1, \sigma), \ldots, \statetf^d(\textsf{v}_d, \sigma) \rangle$ \nllabel{ln:rm-transition}\\
        \For{$i \leq d$\nllabel{ln:loop-reward}}{
            $r_i \leftarrow \rewardtf^i(\textsf{v}_i, \textsf{v}'_i)$ \nllabel{ln:reward-i}
        }
        $\mathbf{r} \leftarrow (r_1(s, a), \ldots, r_d(s, a))$ \nllabel{ln:reward-vector}\\
        $\textit{ND}^{\vec{\textsf{v}}}(s,a) \leftarrow \textit{ND}(\bigcup_{a'}Q_{set}^{\vec{\textsf{v}'}}(s',a'))$ \nllabel{ln:nd-update}\\
        $\overline{\mathscr{R}}^{\vec{\textsf{v}}}(s,a) \leftarrow \overline{\mathscr{R}}^{\vec{\textsf{v}}}(s,a) + \frac{\mathbf{r} -\overline{\mathscr{R}}^{\vec{\textsf{v}}}(s,a)}{n(s,a)}$ \nllabel{ln:avg-reward}\\
        }
        $s \leftarrow s'$;
        $\vec{\textsf{u}} \leftarrow \langle \statetf^1(\textsf{u}_1, \sigma), \ldots, \statetf^d(\textsf{u}_d, \sigma) \rangle$ \nllabel{ln:advance}\\
    }
}
\caption{PQLRM}
\label{algorithm:pql_qrm}
\end{algorithm}

The algorithm we propose is described in Algorithm~\ref{algorithm:pql_qrm}.
The procedure starts by initializing one set of Q-vectors $Q_{set}^{\vec{\textsf{v}}}(s,a)$ per pair $(s,a)$ and per joint RM state $\vec{\textsf{v}} = \langle \textsf{v}_1, \ldots, \textsf{v}_d \rangle$ (line~\ref{ln:init-qset}). This is the multi-objective analogue of the family of Q-functions $\{q^u\}_{u \in U}$ maintained by QRM: where QRM stores one scalar Q-value per RM state, PQLRM stores, for each configuration of RM states, a whole set of Pareto-dominating value vectors as in PQL.

At the start of every episode (line~\ref{ln:episodes}), the states of both the environment and of the RMs are reset (lines~\ref{ln:init-state} and \ref{ln:init-rm}).
The agent then interacts with the environment until a terminal state is reached (line~\ref{ln:while}), or until a fixed number of steps have been executed. At each step, an action $a$ is selected (line~\ref{ln:select-action}) following PQL's set-evaluation rule applied to the Q-sets $Q_{set}^{\vec{\textsf{u}}}(s,\cdot)$ (e.g.\ a hypervolume-based or Pareto-cardinality-based $\varepsilon$-greedy strategy). 

The key ingredient borrowed from QRM is the update of all RM states, done at lines~\ref{ln:loop-v}--\ref{ln:avg-reward}: instead of updating only the entry corresponding to the currently visited RM tuple $\vec{\textsf{u}}$, the algorithm iterates over \emph{every} joint RM state $\vec{\textsf{v}}$ and updates $Q_{set}^{\vec{\textsf{v}}}(s,a)$ as if the agent had been in $\vec{\textsf{v}}$ when taking $(s,a)$. For each such $\vec{\textsf{v}}$, the corresponding next RM tuple $\vec{\textsf{v}'}$ is computed from the label $\sigma = L(s')$ of the observed successor state (lines~\ref{ln:label}--\ref{ln:rm-transition}), and the vector reward $\mathbf{r}$ that \emph{would} have been emitted from $\vec{\textsf{v}}$ is synthesized by querying each RM's reward function $\rewardtf^i(\textsf{v}_i, \textsf{v}'_i)$ (lines~\ref{ln:loop-reward}--\ref{ln:reward-vector}). Because the dynamics of the environment are shared across all RM tuples, a single transition $(s,a,s')$ thus produces $|U^1| \cdots |U^d|$ simultaneous updates, drastically improving sample efficiency compared to a naive PQL applied to the cross-product MDP.

The two updates that follow implement the idea of PQL of decoupling the learning of the immediate reward and the future reward.
Line~\ref{ln:nd-update} updates the set $\textit{ND}^{\vec{\textsf{v}}}(s,a)$ of non-dominated future return vectors reachable from $(s,a)$ under joint RM state $\vec{\textsf{v}}$, by collecting the $Q_{set}^{\vec{\textsf{v}'}}(s',a')$ over all potential future actions $a'$ and preserving only the non-dominated vectors. Line~\ref{ln:avg-reward} updates the average immediate reward vectors with the synthesized $\mathbf{r}$ and $n(s,a)$, tracking the number of times action $a$ was taken from state $s$ during training. As in PQL, $Q_{set}^{\vec{\textsf{v}}}(s,a)$ is then reconstructed on the fly via Equation~\ref{eq:pql_qsets}.

Finally, line~\ref{ln:advance} advances both the environment state and the \emph{actually visited} joint RM state $\vec{\textsf{u}}$ according to the labelling of $s'$, which is required for action selection. 

\paragraph{Policy reconstruction}

A concrete policy is extracted by first committing to a target Q-vector $\mathbf{v}^\star \in Q_{set}^{\vec{\textsf{u}}_0}(s_0, a_0)$.

Then, at each step, the joint RM state $\vec{\textsf{u}}$ is advanced alongside $s$. The residual target is updated via $\mathbf{v}^\star \leftarrow (\mathbf{v}^\star - \mathbf{r})/\gamma$, and we then select the next action whose Q-set
contains a vector closest to it. Indexing the Q-set by $\vec{\textsf{u}}$ (rather than $s$ alone) is essential here, as each $Q_{set}^{\vec{\textsf{v}}}(s,a)$ is only meaningful under the assumption that the RMs are in state $\vec{\textsf{v}}$ in the current environment in which the agent is evolving.







\section{Experimental Evaluation}

\begin{figure}
    \centering
    \begin{subfigure}[b]{0.38\columnwidth}
        \centering
        \includegraphics[width=\columnwidth]{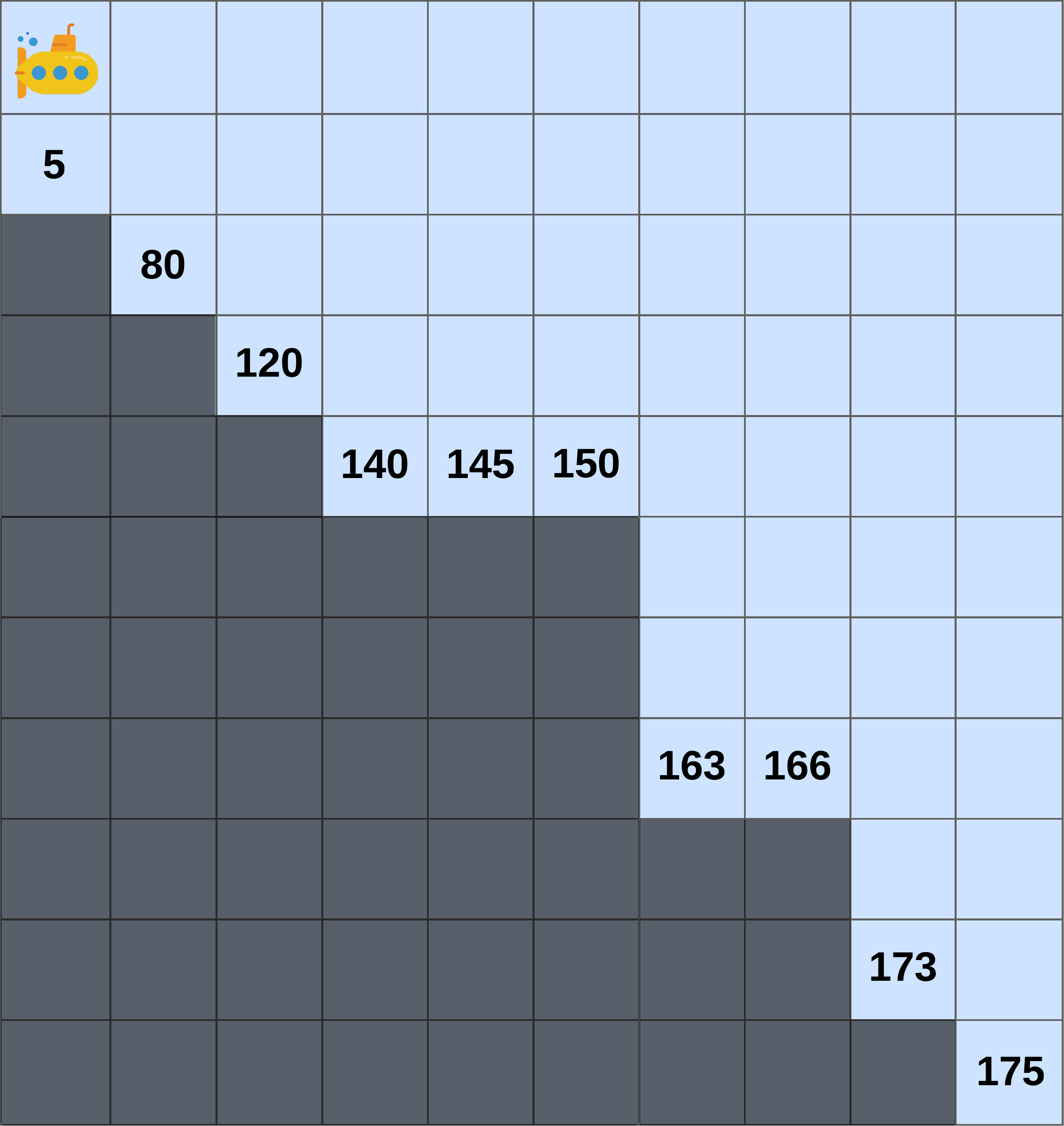}
        \caption{PBST}
        \label{fig:PBST}
    \end{subfigure}
    \hfill
    \begin{subfigure}[b]{0.58\columnwidth}
        \centering
        \begin{tikzpicture}[scale=0.3, font=\scriptsize]
            \draw[gray!55, thin] (0,0) grid (12,9);
            \draw[line width=0.8pt] (0,0) rectangle (12,9);
            \foreach \x in {3,6,9} {
                \draw[line width=0.8pt] (\x,0) -- (\x,1);
                \draw[line width=0.8pt] (\x,2) -- (\x,7);
                \draw[line width=0.8pt] (\x,8) -- (\x,9);
            }
            \draw[line width=0.8pt] (0,6) -- (1,6);
            \draw[line width=0.8pt] (2,6) -- (4,6);
            \draw[line width=0.8pt] (5,6) -- (7,6);
            \draw[line width=0.8pt] (8,6) -- (10,6);
            \draw[line width=0.8pt] (11,6) -- (12,6);
            \draw[line width=0.8pt] (0,3) -- (1,3);
            \draw[line width=0.8pt] (2,3) -- (10,3);
            \draw[line width=0.8pt] (11,3) -- (12,3);
            \node at (1.5,7.5) {D};
            \node at (4.5,7.5) {$\star$};
            \node at (7.5,7.5) {$\star$};
            \node at (10.5,7.5) {C};
            \node at (3.5,8.5) {c};
            \node at (1.5,4.5) {$\star$};
            \node at (4.5,4.5) {o};
            \node at (7.5,4.5) {m};
            \node at (10.5,4.5) {$\star$};
            \node at (8.5,2.5) {c};
            \node at (1.5,1.5) {A};
            \fill[violet!70!blue, draw=black, line width=0.3pt]
                (2.2,1.2) -- (2.8,1.2) -- (2.5,1.85) -- cycle;
            \node at (4.5,1.5) {$\star$};
            \node at (7.5,1.5) {$\star$};
            \node at (10.5,1.5) {B};
        \end{tikzpicture}
        \caption{Office World}
        \label{fig:office}
    \end{subfigure}
    \caption{The two environments used in our experiments. (\subref{fig:PBST}) In PBST, the submarine starts at the top-left corner and must choose among treasures of increasing value located deeper in the grid; dark cells are inaccessible. (\subref{fig:office}) In Office World~\cite{toroicarte2018using}, the agent (triangle) navigates an office-like grid containing several points of interest.}
    \label{fig:environments}
\end{figure}
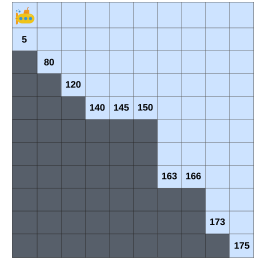
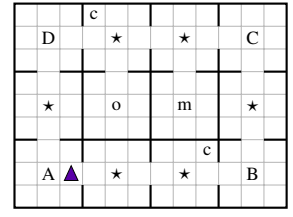

\begin{figure*}[ht]
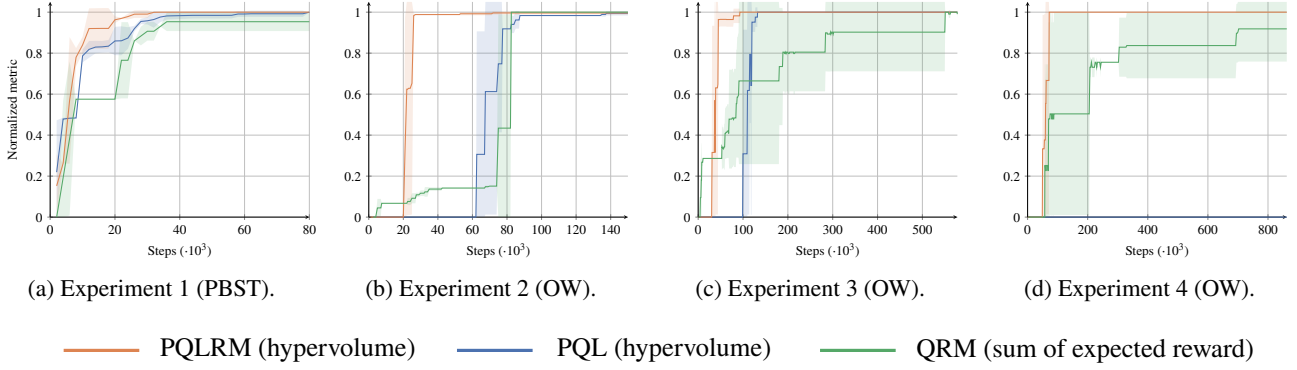

    \begin{subfigure}{0.24\textwidth}
        \center
        \scalebox{0.5}{
\definecolor{expcolor0}{RGB}{76,114,176}
\definecolor{expcolor1}{RGB}{221,132,82}
\definecolor{expcolor2}{RGB}{85,168,104}
\begin{tikzpicture}
\begin{axis}[
xlabel={Steps ($\cdot 10^3$)},
ylabel={Normalized metric},
axis lines=left,
xmin=0,
grid=both,
ymin=0,
ymax=1.05,
]
\addplot[name path=pql_time_treasure_pressurelow, draw=none] coordinates {(2,0) (4,0.4718393013) (6,0.481049332) (8,0.4831836599) (10,0.7549219598) (12,0.7772151349) (14,0.7989251924) (16,0.801966504) (18,0.8038727359) (20,0.790322463) (22,0.7898842154) (24,0.8211653823) (26,0.9059069851) (28,0.9416711621) (30,0.9402204012) (32,0.9240807257) (34,0.9514413235) (36,0.9638888252) (38,0.9653801251) (40,0.966707863) (42,0.9678052173) (44,0.9690638583) (46,0.969219974) (48,0.9693390684) (50,0.9690418451) (52,0.9690422466) (54,0.9693402943) (56,0.9694865135) (58,0.9770142025) (60,0.9770143174) (62,0.9771919393) (64,0.9771919393) (66,0.9771919393) (68,0.9771919393) (70,0.9771919393) (72,0.9771919393) (74,0.9771919393) (76,0.9771919393) (78,0.9771919393) (80,1)};
\addplot[name path=pql_time_treasure_pressureup, draw=none] coordinates {(2,0.4724513305) (4,0.4850907072) (6,0.4843242667) (8,0.4840537759) (10,0.8182702873) (12,0.8595229112) (14,0.8604163565) (16,0.8594837871) (18,0.8641595278) (20,0.9285714662) (22,0.9308572848) (24,0.9265051224) (26,0.9335452691) (28,0.9675243586) (30,0.9755734581) (32,1.004186094) (34,1.000924075) (36,0.9993913011) (38,0.9994264675) (40,0.9994195805) (42,0.999673311) (44,0.9996898373) (46,0.9996907476) (48,0.9996908041) (50,0.9996919645) (52,0.9996919774) (54,0.9996908253) (56,0.999691783) (58,1.005134932) (60,1.005135635) (62,1.007397209) (64,1.007397209) (66,1.007397209) (68,1.007397209) (70,1.007397209) (72,1.007397209) (74,1.007397209) (76,1.007397209) (78,1.007397209) (80,1)};
\addplot[draw=none, fill=expcolor0, fill opacity=0.15] fill between[of=pql_time_treasure_pressurelow and pql_time_treasure_pressureup];
\addplot[thick, draw=expcolor0] coordinates {(2,0.2193549667) (4,0.4784650042) (6,0.4826867993) (8,0.4836187179) (10,0.7865961236) (12,0.818369023) (14,0.8296707744) (16,0.8307251455) (18,0.8340161318) (20,0.8594469646) (22,0.8603707501) (24,0.8738352524) (26,0.9197261271) (28,0.9545977603) (30,0.9578969297) (32,0.9641334097) (34,0.9761826991) (36,0.9816400631) (38,0.9824032963) (40,0.9830637218) (42,0.9837392641) (44,0.9843768478) (46,0.9844553608) (48,0.9845149362) (50,0.9843669048) (52,0.984367112) (54,0.9845155598) (56,0.9845891482) (58,0.9910745672) (60,0.9910749764) (62,0.9922945741) (64,0.9922945741) (66,0.9922945741) (68,0.9922945741) (70,0.9922945741) (72,0.9922945741) (74,0.9922945741) (76,0.9922945741) (78,0.9922945741) (80,1)};
\addplot[name path=pqlrm_time_treasure_pressurelow, draw=none] coordinates {(2,0) (4,0.2622451881) (6,0.260481927) (8,0.7351110258) (10,0.799888417) (12,0.821386397) (14,0.821397866) (16,0.8214069863) (18,0.8214069863) (20,0.9463221175) (22,0.9661492251) (24,0.9646190403) (26,0.9717802249) (28,0.9717933534) (30,0.9717933534) (32,0.9980787597) (34,1) (36,1) (38,1) (40,1) (42,1) (44,1) (46,1) (48,1) (50,1) (52,1) (54,1) (56,1) (58,1) (60,1) (62,1) (64,1) (66,1) (68,1) (70,1) (72,1) (74,1) (76,1) (78,1) (80,1)};
\addplot[name path=pqlrm_time_treasure_pressureup, draw=none] coordinates {(2,0.3064119714) (4,0.2622451881) (6,0.8783198588) (8,0.8249316355) (10,0.8794785945) (12,1.018711221) (14,1.01934941) (16,1.020062516) (18,1.020062516) (20,0.9795495009) (22,0.9731190876) (24,0.9923577043) (26,1.008558946) (28,1.009148102) (30,1.009148102) (32,1.000623105) (34,1) (36,1) (38,1) (40,1) (42,1) (44,1) (46,1) (48,1) (50,1) (52,1) (54,1) (56,1) (58,1) (60,1) (62,1) (64,1) (66,1) (68,1) (70,1) (72,1) (74,1) (76,1) (78,1) (80,1)};
\addplot[draw=none, fill=expcolor1, fill opacity=0.15] fill between[of=pqlrm_time_treasure_pressurelow and pqlrm_time_treasure_pressureup];
\addplot[thick, draw=expcolor1] coordinates {(2,0.1523248672) (4,0.2622451881) (6,0.5694008929) (8,0.7800213306) (10,0.8396835057) (12,0.9200488091) (14,0.9203736378) (16,0.9207347513) (18,0.9207347513) (20,0.9629358092) (22,0.9696341564) (24,0.9784883723) (26,0.9901695855) (28,0.9904707275) (30,0.9904707275) (32,0.9993509323) (34,1) (36,1) (38,1) (40,1) (42,1) (44,1) (46,1) (48,1) (50,1) (52,1) (54,1) (56,1) (58,1) (60,1) (62,1) (64,1) (66,1) (68,1) (70,1) (72,1) (74,1) (76,1) (78,1) (80,1)};
\addplot[name path=qrm_time_treasure_pressurelow, draw=none] coordinates {(2,0) (4,0) (6,0.00767174667) (8,0.5753810003) (10,0.5753810003) (12,0.5753810003) (14,0.5753810003) (16,0.5753810003) (18,0.5753810003) (20,0.5753810003) (22,0.5791784107) (24,0.5791784107) (26,0.8601867815) (28,0.8377939508) (30,0.8611198161) (32,0.8611198161) (34,0.9301643773) (36,0.907816978) (38,0.907816978) (40,0.907816978) (42,0.907816978) (44,0.907816978) (46,0.907816978) (48,0.907816978) (50,0.907816978) (52,0.907816978) (54,0.907816978) (56,0.907816978) (58,0.907816978) (60,0.907816978) (62,0.907816978) (64,0.907816978) (66,0.907816978) (68,0.907816978) (70,0.907816978) (72,0.907816978) (74,0.907816978) (76,0.907816978) (78,0.907816978) (80,0.907816978)};
\addplot[name path=qrm_time_treasure_pressureup, draw=none] coordinates {(2,0) (4,0.5677092536) (6,0.7595029203) (8,0.5753810003) (10,0.5753810003) (12,0.5753810003) (14,0.5753810003) (16,0.5753810003) (18,0.5753810003) (20,0.5753810003) (22,0.9513246314) (24,0.9513246314) (26,0.8601867815) (28,0.9292313427) (30,0.9525572079) (32,0.9525572079) (34,0.9301643773) (36,0.9990688584) (38,0.9990688584) (40,0.9990688584) (42,0.9990688584) (44,0.9990688584) (46,0.9990688584) (48,0.9990688584) (50,0.9990688584) (52,0.9990688584) (54,0.9990688584) (56,0.9990688584) (58,0.9990688584) (60,0.9990688584) (62,0.9990688584) (64,0.9990688584) (66,0.9990688584) (68,0.9990688584) (70,0.9990688584) (72,0.9990688584) (74,0.9990688584) (76,0.9990688584) (78,0.9990688584) (80,0.9990688584)};
\addplot[draw=none, fill=expcolor2, fill opacity=0.15] fill between[of=qrm_time_treasure_pressurelow and qrm_time_treasure_pressureup];
\addplot[thick, draw=expcolor2] coordinates {(2,0) (4,0.1917936668) (6,0.3835873335) (8,0.5753810003) (10,0.5753810003) (12,0.5753810003) (14,0.5753810003) (16,0.5753810003) (18,0.5753810003) (20,0.5753810003) (22,0.7652515211) (24,0.7652515211) (26,0.8601867815) (28,0.8835126467) (30,0.906838512) (32,0.906838512) (34,0.9301643773) (36,0.9534429182) (38,0.9534429182) (40,0.9534429182) (42,0.9534429182) (44,0.9534429182) (46,0.9534429182) (48,0.9534429182) (50,0.9534429182) (52,0.9534429182) (54,0.9534429182) (56,0.9534429182) (58,0.9534429182) (60,0.9534429182) (62,0.9534429182) (64,0.9534429182) (66,0.9534429182) (68,0.9534429182) (70,0.9534429182) (72,0.9534429182) (74,0.9534429182) (76,0.9534429182) (78,0.9534429182) (80,0.9534429182)};
\end{axis}
\end{tikzpicture}
        }
        \caption{Experiment 1 (PBST).}
        \label{subfig:exp_pbst_1}
    \end{subfigure}
    \begin{subfigure}{0.24\textwidth}
        \center
        \scalebox{0.5}{
        \input{plots/convergence_exp1.tex}
        }
        \caption{Experiment 2 (OW).}
        \label{subfig:exp_ow_1}
    \end{subfigure}
    \begin{subfigure}{0.24\textwidth}
        \center
        \scalebox{0.5}{
        \input{plots/convergence_exp2.tex}
        }
        \caption{Experiment 3 (OW).}
        \label{subfig:exp_ow_2}
    \end{subfigure}
    \begin{subfigure}{0.24\textwidth}
        \center
        \scalebox{0.5}{
        \input{plots/convergence_exp3.tex}
        }
        \caption{Experiment 4 (OW).}
        \label{subfig:exp_ow_3}
    \end{subfigure}
    \vspace{1ex}
    \begin{center}
        \begin{tikzpicture}
        \draw[very thick, draw=expcolor1] (0.00,0.00) -- (1.00,0.00);
        \node[anchor=west] at (1.15,0.00) {PQLRM (hypervolume)};
        \draw[very thick, draw=expcolor0] (5.20,0.00) -- (6.2,0.00);
        \node[anchor=west] at (6.4,0.00) {PQL (hypervolume)};
        \draw[very thick, draw=expcolor2] (10.0,0.00) -- (11.0,0.00);
        \node[anchor=west] at (11.15,0.00) {QRM (sum of expected reward)};
        \end{tikzpicture}
    \end{center}
    \caption{Progress of training over the number of training steps. For multi-policy algorithms (i.e. PQLRM and PQL), we report the normalized hypervolume. For QRM, we report the normalized sum of the expected reward of each policy from the initial state. Policies are evaluated every 2,000 steps. The first experiment reports results on PBST and the others on Office World.}
    \label{fig:convergence}
\end{figure*}

\subsection{Experimental Setup}

\paragraph{Setup} 
All experiments were conducted on an AMD Ryzen 7 PRO 5850U processor, with memory usage capped at 8\,GB and a wall-clock limit of 10 minutes per run. Our implementation is in Python 3.10 and builds on the Gymnasium environment interface~\cite{towers2024gymnasium}. Our code is available at: \url{https://github.com/arnaudlequen/PQLRM}.

\paragraph{Baselines}
We compare against PQL, naively adapted to support RMs, by running it on the cross-product state space $\stateset{} \times U^1 \times \cdots \times U^d$, thus folding the joint RM state into the environment. 
We also compare against QRM. The algorithm, however, only learns policies on the extremities of the Pareto-front, since it only searches simultaneously policies that are single-objective.

\subsection{Environments}


\paragraph{Pressurized Bountiful Sea Treasure} In the     
Pressurized Bountiful Sea Treasure (PBST) environment~\cite{moffaert2014multi}, illustrated in Figure~\ref{fig:PBST}, is a grid-world multi-objective reinforcement learning problem in which an agent controls a submarine exploring the seabed to collect treasures. 
The agent must simultaneously optimize three conflicting objectives: minimizing the time to reach a treasure, maximizing the value of the collected treasure, and minimizing the pressure incurred by diving deep underwater. Higher-value treasures are located deeper and farther away, forcing the agent to trade off between speed, safety (low pressure), and reward.

The PBST environment is modeled as a MOMDP: $\mathcal{M} = \langle\mathcal{S}, \mathcal{A}, \mathbf{R}, p\rangle$.  
Episodes terminate when the agent reaches a treasure state. 
A state corresponds to the position of the submarine on a two-dimensional grid: $s = (x, y) \in S$
where $x$ denotes the depth (row index) and $y$ the horizontal position (column index). The grid is of finite size $10 \times 11$. The initial state is located at the surface, at the top-left corner.
In this environment, the black positions are not part of the state space; the positions marked with a number are terminal states containing a treasure whose value corresponds to the number; and the blue positions are part of the state space (see Figure~\ref{fig:PBST}).  
The agent can take four deterministic actions: $\mathcal{A} = \{\emph{up}, \emph{down}, \emph{left}, \emph{right}\}$.
Each action moves the agent to an adjacent cell in the corresponding direction. Actions that would move the agent outside the grid leave the state unchanged.
The transition function is deterministic and defined as:
$$
P(s' \mid s, a) =
\begin{cases}
1 & \text{if } s' = s + \delta(a) \\
0 & \text{otherwise}
\end{cases}
$$
where $\delta(a)$ is the move associated with action $a$. The resulting state is clipped to remain within grid boundaries and reachable positions.
The reward function is vector-valued with three components: $\mathbf{R}(s,a,s') = \big(r_{\text{time}},\; r_{\text{treasure}},\; r_{\text{pressure}}\big)$. 
At each time step, the agent incurs a constant penalty:
$r_{\text{time}} = -1$. 
A positive reward is obtained only when reaching a treasure state:
$$
r_{\text{treasure}} =
\begin{cases}
tr_i & \text{if } s' \text{ is a treasure location } i \\
0 & \text{otherwise}
\end{cases}
$$
where $tr_i$ is the value of treasure located at position $i$; it increases with depth and distance. 
The pressure penalty depends on the number of consecutive down actions performed by the agent. Let $n_{\downarrow}$ denote the number of consecutive \emph{down} actions executed up to and including the current action since the last non-\emph{down} action. Then

$$
r_{\text{pressure}} =
\begin{cases}
-1 & \text{if } n_{\downarrow}=1,\\
-3 & \text{if } n_{\downarrow}=2,\\
-5 & \text{if } n_{\downarrow}\geq 3.
\end{cases}
$$

This reward component is non-Markovian because its value depends on the history of previously executed actions. In particular, the tuple $(s,a,s')$ is not sufficient to determine $r_{\text{pressure}}$, since the same transition may receive different pressure penalties depending on how many consecutive \emph{down} actions preceded it. Consequently, this objective cannot be represented by a standard state-transition reward function and instead requires an RM. The RM tracks the number of consecutive downward movements through its internal state and relies on a propositional symbol emitted whenever a \emph{down} action is performed.
Note that in our experiments, $r_{\text{time}}$ and $r_{\text{treasure}}$ are also encoded by RMs, although this is not a requirement, unlike $r_{\text{pressure}}$.

\subsubsection{Office World}

In the Office World environment~\cite{toroicarte2018using}, an agent navigates an office-like map to complete temporally extended tasks such as delivering coffee or mail, patrolling rooms, or avoiding decorations~(Figure \ref{fig:office}).

The Office World environment is modeled as an MOMDP:
$\mathcal{M} = \langle \mathcal{S}, \mathcal{A}, \mathbf{R}, P \rangle$.
Episodes terminate when the task specification is completed or when the agent violates a forbidden condition (e.g., stepping onto a decoration tile).
A state corresponds to the position of the agent on a two-dimensional grid:
$s = (x,y) \in \mathcal{S}$,
where $x$ and $y$ denote the row and column indices.
The grid contains several special locations associated with propositional events:
coffee locations $c$, mail locations $m$, the office $o$, decorations $\star$, and marked locations $A,B,C,D$ used for patrol tasks.
The agent can execute four deterministic actions:
$\mathcal{A} = \{\emph{up},\emph{down},\emph{left},\emph{right}\}.$
Each action moves the agent to the adjacent grid cell in the corresponding direction.
Actions that would move the agent into a wall or outside the map leave the state unchanged.
The transition dynamics are deterministic and defined by:
$$
P(s' \mid s,a)=
    \begin{cases}
        1 & \text{if } s' = s + \delta(a),\\
        0 & \text{otherwise},
    \end{cases}
$$
where $\delta(a)$ denotes the move induced by action $a$.
The resulting position is constrained to valid cells of the office map.
Tasks are encoded by RMs. For the \emph{get mail} and \emph{get coffee} tasks, the agent receives a reward of $1$ when it reaches the object's location and $1$ when it reaches the office. For the \emph{patrol} task involving going to positions A, B, C, and D (in this order), and the \emph{no hit decoration} task involving not reaching the positions of the decorations, the agent receives a reward only when it has completed its task by reaching the office.

\subsection{Results}


 
We report the convergence speed with regard to the number of iterations of each algorithm in Figure~\ref{fig:convergence}. For multi-policy algorithms (i.e. PQLRM and PQL), we report the normalized hypervolume, while we report the normalized sum of the values of each policy for QRM. Each value is reported every 2,000 steps.
Experiment 1 (Fig.~\ref{subfig:exp_pbst_1}) is performed on the PBST environment, with the three objectives described.
Both PQL and PQLRM reach all 15 Pareto optimal policies. Note that for some treasure there exist two or three policies leading due to the conflicting objectives.
QRM is able to solve the three tasks, but it struggles to reach the farthest treasures. QRM converges a bit slower than the multi-policy algorithms, PQLRM converges slightly faster than PQL.

Experiments 2-4 are performed on the office world environment. In experiment 2 (Fig.~\ref{subfig:exp_ow_1}) the agent learns three tasks to reach the office, to get a coffee and to get the mail in a minimum number of steps. The results show that PQLRM converges faster towards the six Pareto optimal vectors than PQL. QRM is able to solve each task independently.
Experiment 3 (Fig.~\ref{subfig:exp_ow_2}) is performed with the following three objectives: reaching the office without hitting a decoration, reaching the office with a coffee cup and reaching the office with the mail. The results highlight that PQLRM is able to reach all four Pareto optimal policies, while PQL is not able to solve the tasks without hitting decorations and only returns three non-dominated policies that are not Pareto optimal. Moreover, the results show that QRM is not robust on solving all tasks independently, and sometimes it fails to solve the mail task in the number of steps allowed. Finally, experiment 4 (Fig.~\ref{subfig:exp_ow_3}) is performed with two tasks: reaching the office after patrolling all checkpoints (in the right order A, B, C, D) and reaching the office without hitting any decoration. In this case, PQLRM is able to reach the two Pareto optimal policies quite quickly, although the policies are harder to find than for the other experiments. PQL is not able to solve any of the tasks given (i.e. the hypervolume is 0 for it). QRM is able to solve both tasks independently if enough steps are provided.  




\section{Conclusion}

We introduced PQLRM, a multi-objective reinforcement learning algorithm that lifts Pareto Q-Learning to tasks whose objectives are specified by reward machines, by reusing QRM's update method across the joint RM state space.
Experimental results show that PQLRM is more sample-efficient than PQL, while being more flexible than QRM as it synthesizes more Pareto-optimal policies.
Natural next steps include scaling the approach to stochastic environments.

\bibliography{aaai2026}

\end{document}